%% file: Main.tex
\documentclass[sigconf]{acmart}

\AtBeginDocument{%
  }

\usepackage{hyperref}   
\usepackage{url}    
\usepackage{graphicx,color}
\usepackage{bm}
\usepackage{algorithm}
\usepackage{algorithmic}
\usepackage{subfigure}
\usepackage{multirow}
\usepackage{calc}
\usepackage{enumitem,calc}
\usepackage{lipsum}
\input{math_commands}

\newtheorem{problem}{Problem}
\newtheorem{theorem}{Theorem}

\newtheorem{corollary}{Corollary}
\newcommand\preitem{\mdseries\textbullet\space}
\newlist{desclist}{description}{3}
\setlist[desclist,1]{format=\preitem\bfseries,leftmargin=\widthof{\preitem},style=sameline}
\usepackage{expl3}
\ExplSyntaxOn
\newcommand\latinabbrev[1]{
  \peek_meaning:NTF . {
    #1\@}%
  { \peek_catcode:NTF a {
      #1.\@ }%
    {#1.\@}}}
\ExplSyntaxOff

\def\eg{\latinabbrev{e.g}}

\def\ie{\latinabbrev{i.e}}
\newcommand{\name}{{\textsc{MentorGNN}}}

\copyrightyear{2022}
\acmYear{2022}
\setcopyright{acmcopyright}\acmConference[CIKM '22]{Proceedings of the 31st ACM
International Conference on Information and Knowledge Management}{October
17--21, 2022}{Atlanta, GA, USA}
\acmBooktitle{Proceedings of the 31st ACM International Conference on Information
and Knowledge Management (CIKM '22), October 17--21, 2022, Atlanta, GA, USA}
\acmPrice{15.00}
\acmDOI{10.1145/3511808.3557393}
\acmISBN{978-1-4503-9236-5/22/10}


\begin{document}

\title{\name: Deriving Curriculum for Pre-Training GNNs}

\author{Dawei Zhou}
\authornote{Both authors contributed equally to this work.}
\affiliation{%
  \institution{Virginia Tech}
  \country{Blacksburg, Virginia, USA}}
\email{zhoud@vt.edu}

\author{Lecheng Zheng}
\authornotemark[1]
\affiliation{%
  \institution{University of Illinois at Urbana-Champaign}
  \country{Urbana, Illinois, USA}}
\email{lecheng4@illinois.edu}

\author{Dongqi Fu}
\affiliation{%
  \institution{University of Illinois at Urbana-Champaign}
  \country{Urbana, Illinois, USA}}
\email{dongqif2@illinois.edu}

\author{Jiawei Han}
\affiliation{%
  \institution{University of Illinois at Urbana-Champaign}
  \country{Urbana, Illinois, USA}}
\email{hanj@illinois.edu}

\author{Jingrui He}
\affiliation{%
  \institution{University of Illinois at Urbana-Champaign}
  \country{Urbana, Illinois, USA}}
\email{jingrui@illinois.edu}

\renewcommand{\shortauthors}{D. Zhou et al.}

\begin{abstract}
Graph pre-training strategies have been attracting a surge of attention in the graph mining community, due to their flexibility in parameterizing graph neural networks (GNNs) without any label information.
The key idea lies in encoding valuable information into the backbone GNNs, by predicting the masked graph signals extracted from the input graphs. 
In order to balance the importance of diverse graph signals (e.g., nodes, edges, subgraphs), the existing approaches are mostly hand-engineered by introducing hyperparameters to re-weight the importance of graph signals.
However, human interventions with sub-optimal hyperparameters often inject additional bias and deteriorate the generalization performance in the downstream applications.
This paper addresses these limitations from a new perspective, \ie, \emph{deriving curriculum for pre-training GNNs}.
We propose an end-to-end model named \name\ that aims to supervise the pre-training process of GNNs across graphs with diverse structures and disparate feature spaces. 
To comprehend heterogeneous graph signals at different granularities, we propose a curriculum learning paradigm that automatically re-weighs graph signals in order to ensure a good generalization in the target domain.
Moreover, we shed new light on the problem of domain adaption on relational data (\ie, graphs) by deriving a natural and interpretable upper bound on the generalization error of the pre-trained GNNs.
Extensive experiments on a wealth of real graphs validate and verify the performance of \name.
\end{abstract}


\maketitle

\section{Introduction}
\begin{figure*}[t]
\includegraphics[width=0.68\textwidth]{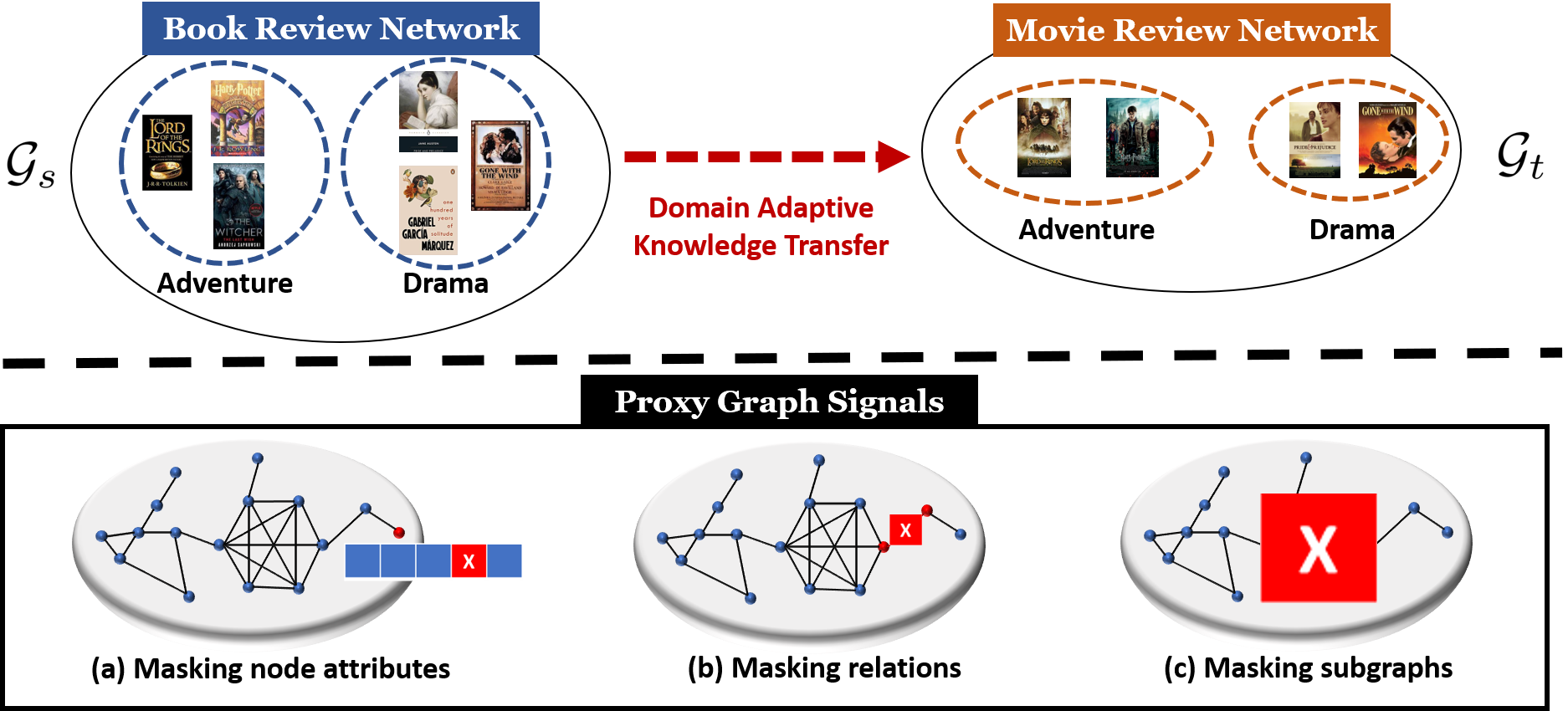}
\centering
\caption{An illustrative example of domain-adaptive graph pre-training. (Top) Domain-adaptive knowledge transfer between the book review network and the movie review network. (Bottom) Proxy graph signals (red masks) at the granularity of nodes, edges, and subgraphs. }
\label{Fig:prb}
\end{figure*}

In the era of big data, graph presents a fundamental data structure for modeling relational data of various domains, ranging from physics~\cite{sanchez2020learning} to chemistry~\cite{DBLP:conf/nips/DuvenaudMABHAA15, DBLP:conf/icml/GilmerSRVD17}, from neuroscience~\cite{de2014graph} to social science~\cite{DBLP:conf/kdd/ZhangT16,DBLP:conf/iclr/GIN}.
Graph neural networks (GNNs)~\cite{DBLP:conf/iclr/GCN,DBLP:conf/iclr/GAT,DBLP:journals/debu/GraphSage,DBLP:conf/iclr/GIN} provide a powerful tool to distill knowledge and learn expressive representations from graph structured data.
Despite the successes of GNNs developments, many GNNs are trained in a supervised manner that requires abundant human-annotated data.
Nevertheless, in many high-impact domains (e.g., brain networks constructed by fMRI)~\cite{guye2010graph,DBLP:conf/kdd/ChoiBSSS17,DBLP:conf/icml/GilmerSRVD17}, collecting high-quality labels is quite resource-intensive and time-consuming. 
To fill this gap, the recent advances~\cite{navarin2018pre,DBLP:conf/iclr/HuLGZLPL20,xu2021infogcl,li2021disentangled} have focused on pre-training GNNs by directly learning from proxy graph signals (shown in Figure~\ref{Fig:prb}) extracted from the input graphs. 
The key assumption is that the extracted graph signals are informative and task-invariant, and as such, the learned graph representation can be generalized well to tasks that have not been observed before. 

Nevertheless, the current GNN pre-training strategies are still at the early stage and suffer from multiple limitations.
Most prominently, the performance of the existing pre-training strategies largely relies on the quality of graph signals. It has been observed that noisy and irrelevant graph signals often lead to negative transfer~\cite{rosenstein2005negativetransfer} and marginal improvement in many application scenarios~\cite{DBLP:conf/nips/HuXQYC0T20}.
For example, one may want to predict the chemical properties~\cite{DBLP:conf/nips/DuvenaudMABHAA15, DBLP:conf/icml/GilmerSRVD17} of a family of novel molecules (\eg, the emerging COVID-19 variants). However, the available data (\eg, the known coronavirus) for pre-training are homologous but with diverse structures and disparate feature spaces. In this case, how can we control the risk of negative transfer and guarantee the generalization performance in the downstream tasks (\eg, characterizing and predicting a new COVID-19 variant)?
Even worse, the risk of negative transfer can be exacerbated when encountering heterogeneous graph signals, i.e., the graph signals can be categorized into distinct types (e.g., class-memberships, attributes, centrality scores, and temporal dependencies) at different granularities (\eg, nodes, edges, and subgraphs)~\cite{DBLP:conf/cikm/FuXLTH20, DBLP:conf/kdd/FuFMTH22, liu2020neural, DBLP:conf/kdd/ZhouZYATDH17, DBLP:journals/tkdd/ZhouZYATDH21}.
To accommodate the heterogeneity of graph signals, the current pre-training strategies are often hand-engineered by using some hyperparameters~\cite{DBLP:conf/iclr/HuLGZLPL20, DBLP:conf/kdd/HuZiniu20}. 
Consequently, the inductive bias from humans might be injected into the pre-trained models thus deteriorating the generalization performance. 

We identify the following two challenges for alleviating the negative transfer phenomenon of pre-training GNNs. 
First (\emph{C1. Cross-Graph Heterogeneity}), how can we distill the informative knowledge from source graphs and translate it effectively to the target graphs for solving novel tasks? 
Second (\emph{C2. Graph-Signal Heterogeneity}), how can we combine and tailor complex graph signals to further harmonize their unique contributions and maximize the overall generalization performance? 
To fill the gap, it is crucial to obtain a versatile GNN pre-training model that can enable knowledge transfer from the source domain to the target domain and carefully exploit the relevant graph signals for downstream tasks. 

In this paper, we propose an end-to-end framework, namely \name, to seamlessly embrace both aforementioned objectives for pre-training GNNs.
In particular, to address C1, we propose a multi-scale encoder-decoder architecture that automatically summarizes graph contextual information at different granularities and learns a mapping function across different graphs.
Moreover, to address C2, we develop a curriculum learning paradigm~\cite{DBLP:conf/icml/MentorNet}, in which a teacher model (\ie, graph signal re-weighting scheme) gradually generates a domain-adaptive curriculum to guide the pre-training process of the student model (\ie, GNNs) and enhance the generalization performance in the tasks of interest. In general, our contributions are summarized as follows.
\begin{desclist}
    \item \textbf{Problem.} We formalize the \emph{domain-adaptive graph pre-training} problem and identify multiple unique challenges inspired by the real applications. 
    \item \textbf{Algorithm.} We propose a novel method named \name, which 1) automatically learns a knowledge transfer function and 2) generates a domain-adaptive curriculum for pre-training GNNs across graphs. In addition, we derive a natural and interpretable generalization bound for domain-adaptive graph pre-training. 
    \item \textbf{Evaluation.} We systematically evaluate the performance of \name\ under two settings: 1) single-source graph transfer and 2) multi-source graph transfer. Extensive results prove the superior performance of \name\ under both settings. 
\end{desclist}

The rest of our paper is organized as follows. In Section 2, we introduce the preliminary and problem definition, followed by the details of our proposed framework \name\ in Section 3. Experimental results are reported in Section 4. We review the related literature in Section 5. Finally, we conclude this paper in Section 6.

\section{Preliminary}
Here we introduce the notations (shown in Table~\ref{TB:Notations}) and the background of our problem setting.
In this paper, we use regular letters to denote scalars (\eg, $\alpha$), boldface lowercase letters to denote vectors (\eg, $\mathbf{v}$), and boldface uppercase letters to denote matrices (\eg, $\mathbf{A}$). 
We denote the source graph and the target graph using $\mathcal{G}_s = (\mathcal{V}_s, \mathcal{E}_s, \mathbf{B}_s)$ and $\mathcal{G}_t = (\mathcal{V}_t, \mathcal{E}_t, \mathbf{B}_t)$, where $\mathcal{V}_s$ ($\mathcal{V}_t$) is the set of nodes, $\mathcal{E}_s$ ($\mathcal{E}_t$) is the set of edges, and $\mathbf{B}_s$ ($\mathbf{B}_t$) is the node attributes in $\mathcal{G}_s$ ($\mathcal{G}_t$). 
Next, we briefly review the graph pre-training strategies and a theoretical model for domain adaptation.

\textbf{Pre-training strategies for GNNs.} Previous studies~\cite{DBLP:journals/corr/KipfW16a, DBLP:conf/nips/HamiltonYL17, navarin2018pre, DBLP:conf/iclr/HuLGZLPL20, DBLP:conf/kdd/HuZiniu20} have been proposed to use easily-accessible graph signals to capture domain-specific knowledge and pre-train GNNs. These attempts are mostly designed to parameterize and optimize the GNNs by predicting the masked graph signals (\eg, node attributes~\cite{DBLP:journals/corr/KipfW16a,DBLP:conf/nips/HamiltonYL17}, edges~\cite{DBLP:conf/www/TangQWZYM15}, subgraphs~\cite{DBLP:conf/iclr/HuLGZLPL20}, and network proximity~\cite{DBLP:conf/kdd/HuZiniu20}) from the visible ones in the given graph $\mathcal{G}$. In Figure~\ref{Fig:prb}, we instantiate the masked graph signals at the level of nodes, edges, and subgraphs, respectively.
To predict a given proxy graph signal $\mathbf{s} \in \mathcal{G}$, we use $f: \mathbf{s}\rightarrow [0,1]$ to denote the true labeling function and $h:\mathbf{s}\rightarrow [0,1]$ to denote the learned hypothesis by GNNs. The risk of the hypothesis $h(\cdot)$ can be computed as $\epsilon(h,f):=\mathbb{E}_{\mathbf{s}\in \mathcal{G}}[|h(\mathbf{s}) - f(\mathbf{s})|]$. In general, the overall learning objective of pre-training models for GNNs can be formulated as 
\begin{align}\label{Obj: spl}\small
    \argmax_{\bm{\theta}} \log h(\mathcal{G}|\hat{\mathcal{G}},\bm{\theta}) = \sum_{\mathbf{s}\in \mathcal{G}} \beta_{\mathbf{s}} \log h(\mathbf{s}|\hat{\mathcal{G}},\bm{\theta})
\end{align}
where $\hat{\mathcal{G}}$ is the corrupted graph with some masked graph signals $\mathbf{s}$, $\beta_{\mathbf{s}}$ is a hyperparameter that balances the weight of the graph signal $\mathbf{s}$, and $\bm{\theta}$ is the hidden parameters of the hypothesis $h(\cdot)$. By maximizing Eq.~\ref{Obj: spl}, we can encode the contextual information of the selected proxy graph signals into pre-trained GNNs $h(\cdot)$. With that, the end-users can fine-tune the pre-trained GNNs and make predictions in the downstream tasks.

\begin{table} [t]
\caption{Symbols and notation.}
\centering
\begin{tabular}{|l|l|}
\hline Symbol&Description\\
\hline
\hline 
$\mathcal{G}_s$, $\mathcal{G}_t$&input source and target graphs.\\
$\mathcal{V}_s$, $\mathcal{V}_t$&the set of nodes in $\mathcal{G}_s$ and $\mathcal{G}_t$.\\
$\mathcal{E}_s$, $\mathcal{E}_t$&the set of edges in $\mathcal{G}_s$ and $\mathcal{G}_t$.\\
$\mathbf{B}_s$, $\mathbf{B}_t$&the node attribute matrices in $\mathcal{G}_s$ and $\mathcal{G}_t$.\\
$\mathbf{X}_s$, $\mathbf{X}_t$&the hidden representation of $\mathcal{G}_s$, $\mathcal{G}_t$.\\
$\mathcal{L}_s$, $\mathcal{L}_t$&the set of annotated data in $\mathcal{G}_s$ and $\mathcal{G}_t$.\\
$n_s$, $n_t$&\# of nodes in $\mathcal{G}_s$ and $\mathcal{G}_t$.\\
$m_s$, $m_t$&\# of edges in $\mathcal{G}_s$ and $\mathcal{G}_t$.\\
\hline 
$\mathcal{G}^{(l)}_s$, $\mathcal{G}^{(l)}_t$&the $l$-th layer coarse graphs of $\mathcal{G}_s$, $\mathcal{G}_t$.\\
$\mathbf{X}^{(l)}_s$, $\mathbf{X}^{(l)}_t$&the hidden representation of $\mathcal{G}^{(l)}_s$, $\mathcal{G}^{(l)}_t$.\\

$\mathbf{A}^{(l)}_s$, $\mathbf{A}^{(l)}_t$&the adjacency matrices of $\mathcal{G}^{(l)}_s$, $\mathcal{G}^{(l)}_t$.\\
$\mathbf{P}^{(l)}_s$, $\mathbf{P}^{(l)}_t$&the perturbation matrices of $\mathcal{G}^{(l)}_s$, $\mathcal{G}^{(l)}_t$.\\
$n^{(l)}_s$, $n^{(l)}_t$&\# of supernodes in $\mathcal{G}^{(l)}_s$, $\mathcal{G}^{(l)}_t$.\\
$L$&\# of layers.\\
$K$&\# of graph signal types.\\
\hline 
$\odot$&Hadamard product.\\
$| \cdot |_p$ & $L_2$-norm of a vector. \\
\hline
\end{tabular}
\label{TB:Notations}
\end{table}

\textbf{Domain adaptation. }
Following the conventional notations in domain adaptation, we let $f_s(\cdot)$ and $f_t(\cdot)$ be the true labeling functions in the source and the target domains. Given $f_s(\cdot)$ and $f_t(\cdot)$, $\epsilon_s (h) = \epsilon_s (h, f_s)$ and $\epsilon_t (h) = \epsilon_t (h, f_t)$ denote the corresponding risks with respect to hypothesis $h(\cdot)$. 
With this, Ben-David et al. proved a domain-adaptive generalization bound in terms of domain discrepancy and empirical risks~\cite{DBLP:journals/ml/Ben-DavidBCKPV10}. To approximate the empirical risks, one common approach~\cite{DBLP:journals/ml/Ben-DavidBCKPV10} is to assume the data points are sampled i.i.d. from both the source and the target domains. 
However, this assumption does not hold for the graph-structured data, as the samples in graphs (nodes, edges, subgraphs) are relational and non-i.i.d. in nature. 
To study the generalization performance of graph mining models, an alternative approach is to define the generalization bound based on the true data distribution~\cite{DBLP:journals/ml/Ben-DavidBCKPV10,DBLP:conf/icml/hanzhao2019}. For instance, Theorem~\ref{theorem: true distribution}~\cite{DBLP:conf/icml/hanzhao2019} provides a population result, which does not rely on the i.i.d. assumption and can be deployed on graphs. 

\begin{theorem}\label{theorem: true distribution}
\normalfont \cite{DBLP:conf/icml/hanzhao2019} 
Let $\left \langle \mathcal{D}_s, f_s \right \rangle$ and $\left \langle\mathcal{D}_t, f_t\right \rangle$ be the source and the target domains, for any function class $\mathcal{H}\subseteq [0,1]^\mathcal{X}$ on the input space $\mathcal{X}$, and $\forall h\in \mathcal{H}$, the following inequality holds:
$$ 
\epsilon_t(h) \leq \epsilon_s(h) + d_{\mathcal{H}}(\mathcal{D}_s, \mathcal{D}_t) + \min \{ \mathbb{E}_{\mathcal{D}_s} [|f_s -f_t |], \mathbb{E}_{\mathcal{D}_t} [|f_s -f_t |]\}
.$$
\end{theorem}

\textbf{Problem definition. }
In this paper, our goal (as shown in Figure~\ref{Fig:prb}) is to learn a knowledge transfer function denoted as $g(\cdot)$, such that the knowledge obtained by a GNN model in the source graph $\mathcal{G}_s$ can be transferred to the target graph $\mathcal{G}_t$ and pre-train the GNNs in $\mathcal{G}_t$. 
Without loss of generality, we assume that 
1) the source graph $\mathcal{G}_s$ is well studied by $f_s(\cdot)$ and comes with rich label information for diverse graph signals $\mathbf{s}\in \mathcal{G}_s$, \ie,
$\mathcal{L}_s =\{(\mathbf{s}, y)| \mathbf{s}\in \mathcal{G}_s, y \in \mathcal{Y}_s\}$; 
and 2) the task of interest in the target graph $\mathcal{G}_t$ is novel, and we are only given a handful of annotated graph signals, \ie, 
$\mathcal{L}_t =\{(\mathbf{s}, y)| \mathbf{s}\in \mathcal{G}_t, y \in \mathcal{Y}_t\}$. 
Given the notations above, we formally define the problem as follows. 

{\setlength{\parindent}{0pt}
\begin{problem}\label{prob}
  \textbf{Domain-Adaptive Graph Pre-Training}\\
	\textbf{Given:} (i) source graph $\mathcal{G}_s = (\mathcal{V}_s, \mathcal{E}_s, \mathbf{B}_s)$ with rich label information $\mathcal{L}_s$,  (ii) target graph $\mathcal{G}_t = (\mathcal{V}_t, \mathcal{E}_t, \mathbf{B}_t)$ with scarce label information $\mathcal{L}_t$, and (iii) user-defined graph neural network architecture.\\
	\textbf{Find:} the pre-trained model $h(\cdot)$ that leverages the knowledge obtained from the source graph $\mathcal{G}_s$ and the target graph $\mathcal{G}_t$.
\end{problem}
}

\section{Proposed Framework \name}
In this section, we introduce our proposed framework \name\ (shown in Figure~\ref{Fig: Framework}) to address Problem 1. The key challenges of Problem 1 lie in the dual-level data heterogeneity, namely \emph{cross-graph heterogeneity} and \emph{graph-signal heterogeneity}. 
Then, we dive into two major modules of \name, \ie, cross-graph adaptation (colored in blue in Figure~\ref{Fig: Framework}) and curriculum learning (colored in orange in Figure~\ref{Fig: Framework}), that are designed specifically for addressing the dual-level data heterogeneity.

\subsection{Cross-Graph Adaptation via Multi-Scale Encoder-Decoder}
The core obstruction of cross-graph adaptation lies in how to effectively translate the knowledge learned from $\mathcal{G}_s$ to $\mathcal{G}_t$ without any supervision of cross-graph association (\eg, partial network alignments~\cite{DBLP:conf/kdd/ZhangT16, yan2021dynamic, yan2021bright, du2021new}).
Specifically, given $\mathcal{G}_s$ and $\mathcal{G}_t$, we want to learn a transformation function $g(\cdot)$ that leverages both network structures and node attributes over the entire graph and translate the knowledge between $\mathcal{G}_s$ and $\mathcal{G}_t$, \ie, $(\mathcal{V}_t, \mathcal{E}_t, \mathbf{B}_t) \simeq g((\mathcal{V}_s, \mathcal{E}_s, \mathbf{B}_s))$. However, learning the transformation function $g(\cdot)$ may require a large parameter space $O(m_sm_t)$ and become computationally intractable, especially when both $\mathcal{G}_s$ and $\mathcal{G}_t$ are large.
In order to alleviate the computational cost, we propose to learn the translation function $g(\cdot)$ at a coarser resolution instead of directly translating knowledge between $\mathcal{G}_s$ and $\mathcal{G}_t$. The underlying intuition is that many real graphs from distinct domains may share similar high-level organizations. 
For instance, in Figure~\ref{Fig:prb} (Top), the book review network ($\mathcal{G}_s$) and the movie review network ($\mathcal{G}_t$) come from two distinct domains, but may share similar high-level organizations (\eg, alignments between book genres and movie genres). That is to say, the communities (the adventure books) on the book review network may have related semantic meanings to the communities (the adventure movies) on the movie review network.

\begin{figure*}[t]
\includegraphics[width=0.66\textwidth]{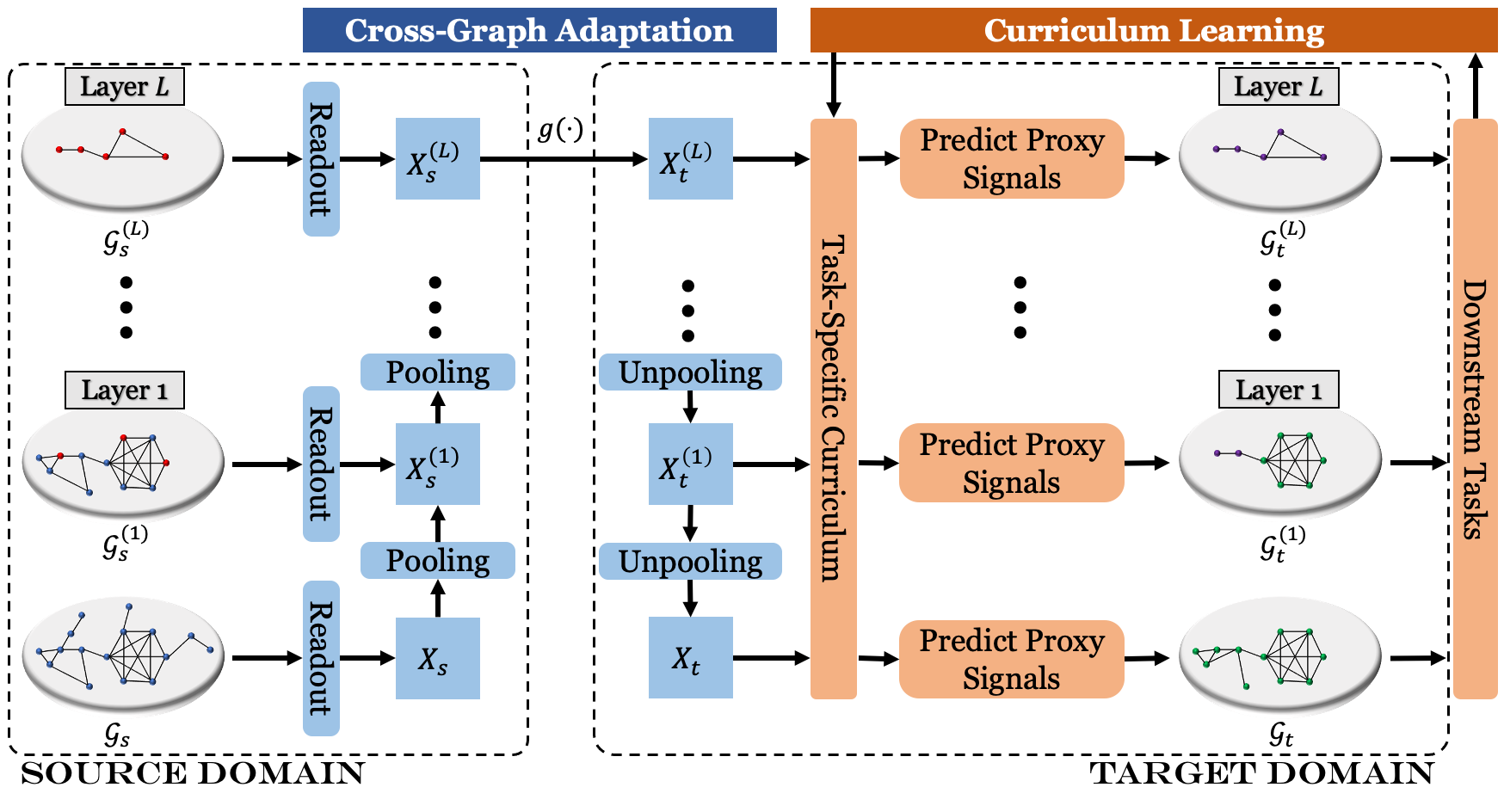}
\centering
\caption{An overview of the \name, which is composed of two modules, namely cross-graph adaptation (colored in blue) and curriculum learning (colored in orange).}
\label{Fig: Framework}
\end{figure*}

Motivated by this observation, we develop a multi-scale encoder-decoder architecture (the module colored in blue in Figure~\ref{Fig: Framework}), which explores the cluster-within-cluster hierarchies of graphs to better characterize the graph signals at multiple granularities. In particular, the encoder $\mathcal{P}$ learns the multi-scale representation of $\mathcal{G}_s$ by pooling the source graph from the fine-grained representation $\mathbf{X}_s$ to the coarse-grained representation $\mathbf{X}_s^{(L)}$, while the decoder $\mathcal{U}$ aims to reconstruct the target graph from the coarse-grained representation $\mathbf{X}_t^{(L)}$ to the fine-grained representation $\mathbf{X}_t$. In our paper, we set an identical number of layers $L$ in the encoder and the decoder to make $\mathcal{G}_s$ and $\mathcal{G}_t$ comparable with each other.

\textbf{Encoder:} The encoder is defined as a set of differentiable pooling matricies $\mathcal{P} = \{{\mathbf{P}}^{(1)}, \ldots, {\mathbf{P}}^{(L)}\}$, where ${\mathbf{P}}^{(l)}\in \mathbb{R}^{{n_s^{(l)}\times n_s^{(l+1)}}}$, $n_s^{(l)}$ and $n_s^{(l+1)}$ are the number of super nodes at layer $l$ and layer $l+1$. Specifically, following~\cite{DBLP:conf/nips/YingY0RHL18}, the differentiable pooling matrix ${\mathbf{P}}^{(l)}$ at layer $l$ is defined as follows.
\begin{align}\small
    \mathbf{P}^{(l)} = \text{softmax}(\text{GNN}_{l,pool} (\mathbf{A}_s^{(l)}, \mathbf{X}_s^{(l)}))
\end{align}
where each entry ${\mathbf{P}}^{(l)} (i,j)$ indicates the assignment coefficient from node $i$ at layer $l$ to the corresponding supernode $j$ at layer $l+1$, and $\text{GNN}_{l,pool}$ is the corresponding surrogate GNN to generate the assignment matrix ${\mathbf{P}}^{(l)}$. 
It is noteworthy that the number of granularity levels $L \geq 2$ and the maximum number of supernodes in each layer $n_s^{(l)}$ are both hyperparameters selected by the end-users. More implementation details are discussed in Section 4.1. 
With the differentiable pooling matrix ${\mathbf{P}}^{(l)}$, the $l^{\text{th}}$-layer coarse-grained representation $ \mathbf{X}_s^{(l)} $ of $\mathcal{G}_s$ can be approximated by $\mathbf{X}_s^{(l)}  = \mathbf{P}^{(l-1)\prime}\ldots \mathbf{P}^{(1)\prime} \mathbf{X}_s$.

\textbf{Decoder:} Different from the encoder, our decoder is composed of a translation function $g(\cdot)$ and a set of differentiable unpooling matrices $\mathcal{U} = \{{\mathbf{U}}^{(1)}, \ldots, {\mathbf{U}}^{(L)}\}$. To be specific, the translation function $g(\cdot)$ learns a non-linear mapping between $\mathcal{G}_s$ and $\mathcal{G}_t$ at the coarsest layer $L$. In our implementation,  $g(\cdot)$ is parameterized by a multilayer perceptron (MLP), and the differentiable pooling matrix ${\mathbf{U}}^{(l)}\in \mathbb{R}^{n_t^{(l+1)}\times n_t^{(l)}}$ at each layer $l$ can be computed as follows.
\begin{align}\small
    \mathbf{U}^{(l)} = \text{softmax}(\text{GNN}_{l,unpool} (\mathbf{A}_t^{(l)}, \mathbf{X}_t^{(l)}))
\end{align}
In contrast to $\text{GNN}_{l,pool}$, $\text{GNN}_{l,unpool}$ is a reverse operation that aims to reconstruct ${\mathbf{X}}^{(l-1)}$ from its coarse representation ${\mathbf{X}}^{(l)}$. With the learned translation function $g(\cdot)$ and the differentiable unpooling matrices $\mathcal{U}$, the hidden representation of $\mathcal{G}_t$ can be computed by $\hat{\mathbf{X}}_t  = \mathbf{U}^{(1)\prime}\ldots \mathbf{U}^{(L)\prime} \hat{\mathbf{X}}_t^{(L)} $, 
where $ \hat{\mathbf{X}}_t^{(L)} = g({\mathbf{X}}_s^{(L)}) $ is the translated representation from $\mathcal{G}_s$ to $\mathcal{G}_t$ at the $L^\text{th}$ layer.

\subsection{Graph Signal Comprehension via Curriculum Learning}\label{section: CL}
In the presence of various types of graph signals, the current GNNs pre-trained models are mostly designed in the form of a weighted combination of multiple graph signal encoders by incorporating some hyperparameters. However, manually selecting hyperparameters are often challenging and may lead to sub-optimal performance due to the inductive bias from humans.
Here we propose a graph signal re-weighting scheme (the module colored in orange in Figure~\ref{Fig: Framework}) to capture the contribution of each graph signal towards the downstream tasks. 
The learned sample weighting scheme specifies a curriculum under which the GNNs will be pre-trained gradually from the easy concepts to the hard concepts. In our problem setting, the curriculum is presented as a sequence of graph signals that are extracted from the given graph.
Compared with the previous GNNs pre-training methods, \name\ automatically 1) assigns an attention weight towards each graph signal without the requirements of the pre-defined hyperparameters, and 2) supervises the pre-training process to control the risk of negative transfer.

Consider a GNN-based pre-training problem with $K$ types of graph signals extracted from graphs at $L$ levels of resolutions, we formulate the learning objective as follows. 
\begin{align}\label{eq:obj_main}\small
    \mathcal{L} = \argmin_{\mathbf{\Theta},\bm{\theta}} \ 
    \sum_{l=1}^L \sum_{k=1}^K 
    \sum_{\mathbf{s}_i^{(l, k)} \in \mathcal{G}_t^{(l)}} &g_m(\mathbf{s}_i^{(l, k)};\mathbf{\Theta})\mathcal{J}(\mathbf{Y}_t, h(\hat{\mathbf{X}}_t^{(l)},\mathbf{s}_i^{(l, k)}, \bm{\theta}))\nonumber\\
    &+ G(g_m(\mathbf{s}_i^{(l, k)}, \mathbf{\Theta}), \bm{\lambda})
\end{align}
where 
$\mathbf{s}_i^{(l,k)}$ denotes the $i^\text{th}$ sample of the type-$k$ graph signals (\eg, edges, node attributes, centrality scores) extracted from the $l^\text{th}$-layer of $\mathcal{G}_t$,
$g_m(\mathbf{s}^{(l,k)}_i;\mathbf{\Theta})$ is a function that predicts the attention weights for each training graph signal $\mathbf{s}^{(l,k)}_i$, 
$h(\hat{\mathbf{X}}_t^{(l)},\mathbf{s}_i^{(l, k)}, \bm{\theta})$ denotes the pre-trained GNNs using $\mathbf{s}_i^{(l, k)}$, $\mathcal{J}(\mathbf{Y}_t, h(\hat{\mathbf{X}}_t^{(l)},\mathbf{s}_i^{(l, k)}, \bm{\theta}))$ denotes the prediction loss of a downstream task over a handful of labeled examples $\mathbf{Y}_t$, and $G(g_m(\mathbf{s}_i^{(l, k)}, \mathbf{\Theta}), \bm{\lambda})$ is the curriculum regularizer parameterized by the learning threshold $\bm{\lambda}$. 
It is worthy to mention that such labeled examples can be any type of graph signals, including the class-memberships of nodes, edges, and even subgraphs. 
The goal of $\mathcal{L}$ is to automatically learn attention weight $g_m(\mathbf{s}^{(l,k)}_i;\mathbf{\Theta})$ towards each graph signal $\mathbf{s}_i^{(l, k)}$ in order to guide the pre-training process of GNNs $h(\cdot)$ via curriculum learning, thus mitigating the risk of negative transfer.

Intuitively, the learning objective in Eq.~\ref{eq:obj_main} can be interpreted as a teacher-student training paradigm~\cite{DBLP:conf/nips/KumarPK10,DBLP:conf/aaai/Zhu15,DBLP:conf/iclr/FanTQ0L18,DBLP:conf/kdd/ZhouHYF18}. In particular, 
the labeling function $h(\hat{\mathbf{X}}_t^{(l)},\mathbf{s}_i^{(l, k)}, \bm{\theta})$ parameterized by $\bm{\theta}$ serves as the student model, which aims to learn expressive representation by predicting the graph signal $\mathbf{s}_i^{(l,k)}$; the teacher model $g_m(\mathbf{s}_i^{(l, k)};\mathbf{\Theta})$ parameterized by $\mathbf{\Theta}$ aims to measure the importance of each graph signal $\mathbf{s}_i^{(l,k)}$, and to provide guidance to pre-train GNNs on the target graph $\mathcal{G}_t$. 
To regularize the learning curriculum, one prevalent choice is to employ some pre-defined curriculums~\cite{DBLP:journals/corr/KipfW16a, DBLP:conf/nips/HamiltonYL17,  navarin2018pre, DBLP:conf/iclr/PetarH19, DBLP:conf/iclr/HuLGZLPL20, DBLP:conf/kdd/HuZiniu20, DBLP:conf/www/ZhouNMFH20, DBLP:conf/kdd/ZhouNH18, wang2021curgraph}, which have been extensively explored in the literature~\cite{DBLP:conf/nips/KumarPK10,DBLP:conf/aaai/Zhu15,DBLP:conf/aaai/JiangMZSH15,DBLP:conf/iclr/FanTQ0L18}. Here we consider a curriculum regularizer derived from the robust non-convex penalties~\cite{DBLP:conf/aaai/JiangMZSH15}. 
\begin{align}\small\nonumber
    G(g_m(\mathbf{s}_i^{(l, k)}, \mathbf{\Theta}), \lambda_1, \lambda_2) =  \frac{1}{2} \lambda_2 g_m^2(\mathbf{s}_i^{(l, k)};\mathbf{\Theta}) - (\lambda_1 + \lambda_2)g_m(\mathbf{s}_i^{(l, k)};\mathbf{\Theta})
\end{align}
where $\bm{\lambda} = \{ \lambda_1, \lambda_2 \} $ are both positive hyperparameters. Note that $G(g_m(\mathbf{s}_i^{(l, k)}, \mathbf{\Theta}), \lambda_1, \lambda_2)$ is a convex function in terms of $g_m(\mathbf{s}_i^{(l, k)}, \mathbf{\Theta})$ and thus the closed-form solution of our learning curriculum can be easily obtained as follows. 
\begin{align}\label{update Theta}
&g_m(\mathbf{s}^{(l,k)}_i; \mathbf{\Theta}^*)\nonumber\\
&=
\begin{cases}
1(\mathcal{J}(\mathbf{Y}_t, h(\hat{\mathbf{X}}_t^{(l)},\mathbf{s}_i^{(l, k)}, \bm{\theta})) \leq \lambda_1) &\text{$\lambda_2=0$}\\
\min(\max(0,1-\frac{\mathcal{J}(\mathbf{Y}_t, h(\hat{\mathbf{X}}_t^{(l)},\mathbf{s}^{(l, k)}, \bm{\theta})) - \lambda_1}{\lambda_2}), 1) & \text{Otherwise}
\end{cases}
\end{align}

\begin{algorithm}[t]
\caption{The \name\ learning framework.}
\label{Alg}
\begin{algorithmic}[1]
\REQUIRE :
    (i) source graph $\mathcal{G}_s = (\mathcal{V}_s, \mathcal{E}_s, \mathbf{B}_s)$ with rich label information $\mathcal{L}_s$, (ii) target graph $\mathcal{G}_t = (\mathcal{V}_t, \mathcal{E}_t, \mathbf{B}_t)$ with scarce label information $\mathcal{L}_t$, (iii) user-defined graph neural network architecture $h(\cdot)$ for pre-training, (iv) parameters $\lambda_1$, $\lambda_2$, and $\xi$.
\ENSURE :
    (i) the knowledge transfer function $g(\cdot)$, (ii) the pre-trained model $h(\cdot)$ that leverages the knowledge obtained from both $\mathcal{G}_s$ and $\mathcal{G}_t$.
    \STATE Initialize $\bm{\theta}$ and $\mathbf{\Theta}$.
    \WHILE{Stopping criterion is not satisfied}
        \STATE Fetch a collection of graph signals $\mathbf{s}_i^{(l, k)}\in \mathcal{G}_t$. 
        \IF{Update teacher}
            \STATE Update $\mathbf{\Theta}$ by taking a gradient step for $\mathcal{L}$.
            \STATE Compute learning curriculum $g_m(\textbf{s}_i^{(l, k)};\mathbf{\Theta})$ via the closed-form solution in Eq.~\ref{update Theta}.
        \ENDIF
        \IF{Update student}
            \STATE Update $\bm{\theta}$ and $g(\cdot)$ by taking a gradient step for $\mathcal{J}(\mathbf{Y}_t, h(\hat{\mathbf{X}}_t^{(l)},\textbf{s}^{(l, k)}, \bm{\theta}))$.
            \STATE Compute $\{\hat{\mathbf{X}}_t^{(1)}, \ldots, \hat{\mathbf{X}}_t^{(L)}\}$ via multi-scale encoder-decoder (Subsection 3.1).
        \ENDIF
        \STATE Augment the values of $\lambda_1$ and $\lambda_2$ by the ratio $\xi$.
    \ENDWHILE
\end{algorithmic}
\end{algorithm}

The learning threshold $\bm{\lambda} = \{\lambda_1, \lambda_2\}$ plays a key role in controlling the learning pace of \name. When $\lambda_2 = 0$, the algorithm will only select the ``easy'' graph signals of $\mathcal{J}(\mathbf{Y}_t, h(\hat{\mathbf{X}}_t^{(l)},\mathbf{s}_i^{(l, k)}, \bm{\theta})) \leq \lambda_1$ in training labeling function $h(\cdot)$, which is close to the binary scheme in self-paced learning~\cite{DBLP:conf/nips/KumarPK10,DBLP:conf/aaai/JiangMZSH15}. When $\lambda_2 \neq 0$, $g_m(\mathbf{s}^{(l,k)}_i; \mathbf{\Theta}^*)$ will return continuous values in $[0,1]$, and the graph signals with the loss of $\mathcal{J}(\mathbf{Y}_t, h(\hat{\mathbf{X}}_t^{(l)},\mathbf{s}_i^{(l, k)}, \bm{\theta})) \leq \lambda_1 + \lambda_2 $ will not be selected in pre-training. In practice, we gradually augment the value of $\lambda_1 + \lambda_2$ to enforce \name\ learning from the ``easy'' graph signals to the ``hard'' graph signals by mimicking the cognitive process of humans and animals~\cite{DBLP:conf/icml/BengioLCW09}. 
The pseudo-code of \name\ is provided in Algorithm~\ref{Alg}, which is designed in an alternative-updating fashion~\cite{tseng2001blockcoordinate}. 
In each iteration, when we train the student model $h(\hat{\mathbf{X}}_t^{(l)},\mathbf{s}_i^{(l, k)}, \bm{\theta}))$, we keep $\mathbf{\Theta}$ fixed and minimize the prediction loss $\mathcal{J}(\mathbf{Y}_t, h(\hat{\mathbf{X}}_t^{(l)},\mathbf{s}^{(l, k)}, \bm{\theta}))$; 
when we train the teacher model $g_m(\mathbf{s}_i^{(l, k)};\mathbf{\Theta})$, we keep $\bm{\theta}$ fixed and update the learning curriculum to be used by the student model in the next iteration; at last, we augment the value of Augment the values of $\lambda_1$ and $\lambda_2$ by the ratio $\xi$ for the next iteration.  


\subsection{Generalization Bound of \name}
Given a source graph $\mathcal{G}_s$ and a target graph $\mathcal{G}_t$, what generalization performance can we achieve with \name\ in pre-training GNNs? Here, we present two approaches towards the generalization bound of \name\ in the presence of multiple types of graph signals at different granularities: one by a union bound argument and the other relying on the graph signal learning curriculum. 
Let $f_s(\cdot)$ be the true labeling function in $\mathcal{G}_s$, and $f_t^{(l,k)}(\cdot)$ be the true labeling function for the $k^{\text{th}}$ type graph signals $\mathbf{s}^{(l,k)}$ at the $l^\text{th}$ level of $\mathcal{G}_t$.
One straightforward idea is to leverage the generalization error between $f_s(\cdot)$ and each $f_t^{(l,k)}(\cdot)$ by applying Theorem~\ref{theorem: true distribution} $l\times k$ times. Following this idea, we can obtain the following worst-case generalization bound of \name, which largely depends on the largest generalization error between $f_s(\cdot)$ and $f_t^{(l,k)}(\cdot)$.

\begin{corollary}\label{theorem: worst case}
\normalfont (Worst Case)
Given $\mathcal{G}_s$ and $\mathcal{G}_t$, $f_s(\cdot)$ is the labeling function in $\mathcal{G}_s$, and $f_t^{(l,k)}(\cdot)$ is the labeling function for the $k^{\text{th}}$ type graph signals $\mathbf{s}^{(l,k)}$ at the $l^\text{th}$ level of granularity in $\mathcal{G}_t$. 
Then, for any function class $\mathcal{H}\subseteq [0,1]^\mathcal{X}$, and $\forall h\in \mathcal{H}$, the following inequality holds: 
\begin{align}\small
\epsilon_t(h) \leq &\epsilon_s(h) + d_{\mathcal{H}}(\mathcal{G}_s, \mathcal{G}_t) \nonumber\\
&+ \max_{l,k} \{\min \{ \mathbb{E}_{\mathcal{G}_s} [|f_s -f_t^{(l,k)} |], \mathbb{E}_{\mathcal{G}_t} [|f_s -f_t^{(l,k)} |]\}\}
\end{align}
\end{corollary}

The worst-case generalization bound shown in Corollary~\ref{theorem: worst case} could be pessimistic in practice, especially when the graphs are large and noisy. However, extensive work~\cite{navarin2018pre,DBLP:conf/iclr/HuLGZLPL20,DBLP:conf/kdd/HuZiniu20} has empirically shown that leveraging multiple types of graph signals often leads to improved performance in many application domains, even in the presence of noisy data and irrelevant features. The key observation is that the information from multiple types of graph signals is often redundant and complementary. That is to say, when the majority of graph signals are related, a few irrelevant graph signals may not hurt the overall generalization performance too much. Hence, the natural question is: \emph{can we obtain a better generalization bound than the one shown in Corollary~\ref{theorem: worst case}?} To answer this question, we present a re-weighting case of the generalization bound for \name, that is developed based on the obtained learning curriculum $g_m(\mathbf{s}^{(l,k)}, \Theta)$ as follows. 
\begin{corollary}\label{theorem: average case}
\normalfont (Re-weighting Case)
Given $\mathcal{G}_s$ and $\mathcal{G}_t$, $f_s(\cdot)$ is the labeling function in $\mathcal{G}_s$, and $f_t^{(l,k)}(\cdot)$ is the labeling function for the $k^{\text{th}}$ type graph signals $\mathbf{s}^{(l,k)}$ at the $l^\text{th}$ level of granularity in $\mathcal{G}_t$.  
Then, for any function class $\mathcal{H}\subseteq [0,1]^\mathcal{X}$, $\forall h\in \mathcal{H}$, and $\sum_{l,k,i} g_m(\mathbf{s}_i^{(l,k)}, \Theta) = 1$, the following inequality holds:
\begin{align}\small
\epsilon_t(h) \leq & \epsilon_s(h) + d_{\mathcal{H}}(\mathcal{G}_s, \mathcal{G}_t)+\sum_{l,k,i} g_m(\mathbf{s}_i^{(l,k)}, \Theta) \nonumber\\
&\cdot \min \{ \mathbb{E}_{\mathcal{G}_s} [|f_s -f_t^{(l,k)} |], \mathbb{E}_{\mathcal{G}_t} [|f_s -f_t^{(l,k)}|]\}\} \nonumber
\end{align}

\end{corollary}


\textbf{Remark 1: }Corollary~\ref{theorem: worst case} and Corollary~\ref{theorem: average case} are both based on the true data distribution, and as such, the derived generalization bound might slightly deviate from the empirical results in practice. However, we argue that the state-of-the-art GNNs~\cite{DBLP:conf/iclr/GCN, DBLP:conf/iclr/GAT, DBLP:conf/iclr/GIN} are reliable and accurate, whose empirical performance is very close to the true labeling function. Therefore, our theoretical results can well approximate the generalization bound of \name\ in practice. 

\textbf{Remark 2: }Corollary~\ref{theorem: average case} reduces the multi-type multi-granularity of graph signals into an aggregated version with a linear combination using $g_m(\mathbf{s}_i^{(l,k)}, \Theta)$. In fact, the worst-case generalization bound can be considered as a special case of the generalization bound shown in Corollary~\ref{theorem: average case}. Based on the following inequality, it is easy to prove that Corollary~\ref{theorem: average case} provides a much tighter generalization bound than Corollary~\ref{theorem: worst case}.
\begin{align}\small
&\sum_{l,k,i} g_m(\mathbf{s}_i^{(l,k)}, \Theta)\min \{ \mathbb{E}_{\mathcal{G}_s} [|f_s -f_t^{(l,k)} |], \mathbb{E}_{\mathcal{G}_t} [|f_s -f_t^{(l,k)}|]\}\}\nonumber\\
&\leq \max_{l,k} \{\min \{ \mathbb{E}_{\mathcal{G}_s} [|f_s -f_t^{(l,k)} |], \mathbb{E}_{\mathcal{G}_t} [|f_s -f_t^{(l,k)} |]\}\}.\nonumber
\end{align}

\begin{table*} [htp]
\small
\caption{Accuracy of node classification in the single-source graph transfer setting, from a single source graphs (Cora) to a single target graph (Reddit1 or  Reddit2 or  Reddit3).
}
\centering
\scalebox{1}{
\begin{tabular}{|c|c|c|c|}
\hline Method & Cora $\rightarrow$ Reddit1 & Cora $\rightarrow$ Reddit2 & Cora $\rightarrow$ Reddit3 \\
\hline
\hline  GCN     & 0.8736 $\pm$ 0.0151 & 0.8996 $\pm$ 0.0158 & 0.8816 $\pm$ 0.0050 \\
\hline  GAT     & 0.9420 $\pm$ 0.0154 & 0.9241 $\pm$ 0.0094 & 0.8985 $\pm$ 0.0088 \\
\hline  DGI     & 0.7845 $\pm$ 0.0208 & 0.9062 $\pm$ 0.0071 & 0.8388 $\pm$ 0.0098 \\
\hline  GPA     & 0.7011 $\pm$ 0.0116 & 0.7157 $\pm$ 0.0053 & 0.7271 $\pm$ 0.0063 \\
\hline UDA-GCN	& 0.9323 $\pm$ 0.0106 & 0.9378 $\pm$ 0.0079	& 0.9528 $\pm$ 0.0058 \\
\hline GPT-GNN	& \textbf{0.9570 $\pm$ 0.0097} & 0.9614 $\pm$ 0.0065	& 0.9548 $\pm$ 0.0092 \\
\hline Pretrain-GNN & 0.9484 $\pm$ 0.0105 &	0.9351 $\pm$ 0.0100 &	0.9574 $\pm$ 0.0090 \\
\hline  \name-V & 0.9448 $\pm$ 0.0131 & 0.9584 $\pm$ 0.0045 & 0.9454 $\pm$ 0.0071 \\
\hline  \name-C & 0.9508 $\pm$ 0.0097 & 0.9640 $\pm$ 0.0060 & 0.9655 $\pm$ 0.0072 \\
\hline  \name   & 0.9562 $\pm$ 0.0059 & \textbf{0.9815 $\pm$ 0.0053} & \textbf{0.9741 $\pm$ 0.0046}\\
\hline
\end{tabular}}
\label{TB:Single_graph_cora}
\end{table*}

\begin{table*} [htp]
\small
\caption{Accuracy of node classification in the single-source graph transfer setting, from a single source graphs (CiteSeer) to a single target graph (Reddit1 or  Reddit2 or  Reddit3).
}
\centering
\scalebox{1}{
\begin{tabular}{|c|c|c|c|}
\hline Method & CiteSeer $\rightarrow$ Reddit1 & CiteSeer $\rightarrow$ Reddit2 & CiteSeer $\rightarrow$ Reddit3 \\
\hline
\hline  GCN     & 0.8736 $\pm$ 0.0151 & 0.8996 $\pm$ 0.0158	& 0.8816 $\pm$ 0.0050 \\
\hline  GAT     & 0.9420 $\pm$ 0.0154 & 0.9241 $\pm$ 0.0094	& 0.8985 $\pm$ 0.0088 \\
\hline  DGI     & 0.7845 $\pm$ 0.0208 & 0.9062 $\pm$ 0.0071	& 0.8388 $\pm$ 0.0098 \\
\hline  GPA     & 0.6932 $\pm$ 0.0105 & 0.7162 $\pm$ 0.0055	& 0.7273 $\pm$ 0.0065 \\
\hline UDA-GCN	& 0.9027 $\pm$ 0.0082 & 0.9540 $\pm$ 0.0092	& 0.9610 $\pm$ 0.0131 \\
\hline GPT-GNN	& 0.9544 $\pm$ 0.0078 & 0.9528 $\pm$ 0.0068	& 0.9541 $\pm$ 0.0100 \\
\hline Pretrain-GNN & 0.9386 $\pm$ 0.0084 & 0.9488 $\pm$ 0.0098	& 0.9581 $\pm$ 0.0113 \\
\hline  \name-V & 0.9589 $\pm$ 0.0043 & 0.9637 $\pm$ 0.0072	& 0.9763 $\pm$ 0.0034 \\
\hline  \name-C & 0.9521 $\pm$ 0.0091 & 0.9646 $\pm$ 0.0064	& 0.9779 $\pm$ 0.0045 \\
\hline  \name   & \textbf{0.9617 $\pm$ 0.0028} & \textbf{0.9834 $\pm$ 0.0048}	& \textbf{0.9806 $\pm$ 0.0026} \\
\hline
\end{tabular}}
\label{TB:Single_graph_citeseer}
\end{table*}

\section{Experiments}
We evaluate the performance of \name\ on both synthetic and real graphs in the following aspects:
\begin{desclist}
\setlength{\belowdisplayskip}{-6pt}
\item[Effectiveness: ]We report comparison results with a diverse set of baseline methods, including state-of-the-art GNNs, graph pre-training models, and transfer learning models, in the single-source graph transfer and the multi-source graph transfer settings.
\item[Case study: ]We conduct a case study to study the generalization performance of \name\ in the dynamic protein-protein interaction graphs.
\item[Parameter sensitivity and scalability analysis: ]We study the parameter sensitivity and the scalability of \name\ on both synthetic and real graphs. 
\end{desclist}

\subsection{Experimental Setup}
\textbf{Data sets:} 
Cora~\cite{DBLP:conf/icml/LuG03} data set is a citation network consisting of 2,708 scientific publications and 5,429 edges. Each edge in the graph represents the citation from one paper to another. 
CiteSeer~\cite{DBLP:conf/icml/LuG03} data set consists of 3,327 scientific publications, which could be categorized into six classes, and this citation network has 9,228 edges. 
PubMed~\cite{namata2012query} is a diabetes data set, which consists of 19,717 scientific publications in three classes and 88,651 edges. 
The Reddit~\cite{hamilton2017loyalty} data set was extracted from Reddit posts in September 2014. After pre-processing, three Reddit graphs are extracted and denoted as Reddit1, Reddit2, and Reddit3, respectively. Reddit1 consists of 4,584 nodes and 19,460 edges; Reddit2 consists of 3,765 nodes and 30,494 edges; and Reddit3 consists of 4,329 nodes and 35,191 edges.\\
\textbf{Baselines:} We compare our method with the following baselines: 
(1) GCN~\cite{DBLP:conf/iclr/GCN}: graph convolutional network, which is directly trained and tested on the target graph; 
(2) GAT~\cite{DBLP:conf/iclr/GAT}: graph attention network, which is directly trained and tested on the target graph; 
(3) DGI~\cite{DBLP:conf/iclr/PetarH19}: deep graph infomax, which is directly trained and tested on the target graph; 
(4) GPA~\cite{DBLP:conf/nips/HuXQYC0T20}: a policy network for transfer learning across graphs. Since GPA is designed for zero-shot setting, we fine-tune the pre-trained model with 100 iterations in the downstream tasks, and then make the final prediction; 
(5) UDA-GCN~\cite{DBLP:conf/www/WuP0CZ20}: an unsupervised domain-adaptive graph convolutional network; 
(6) GPT-GNN~\cite{DBLP:conf/kdd/HuZiniu20}: a self-supervised attributed graph generative framework to pre-train a GNN by capturing both the structural and semantic properties of the graph;
(7) Pretrain-GNN~\cite{DBLP:conf/iclr/HuLGZLPL20}: a self-supervised pre-training model that learns both local and global representations of each node;
(8) \name-V: one variant of \name, which only considers node attributes as graph signals; 
and (9) \name-C: one variant of \name, which is designed without the curriculum learning module. \\
\textbf{Implementation details:} The data and code are available in the link\footnote{\url{https://github.com/Leo02016/MentorGNN}}. In the implementation of \name, we consider two types ($K =2$) of graph signals, \ie, node attributes and edges, at $L=3$ levels of granularity. The output dimension of the first level of granularity is 500, and the output dimension of the second level of granularity is 100. We use Adam~\cite{DBLP:journals/corr/KingmaB14} as the optimizer with a learning rate of 0.005 and a two-layer GAT~\cite{DBLP:conf/iclr/GAT} with a hidden layer of size 50 as our backbone structure. \name\ and its variants are trained for a maximum of 2000 episodes. The experiments are performed on a Windows machine with eight 3.8GHz Intel Cores and a single 16GB RTX 5000 GPU.

\begin{table*} [htp]
\small
\caption{Accuracy of node classification in the single-source graph transfer setting, from a single source graphs (PubMed) to a single target graph (Reddit1 or  Reddit2 or  Reddit3).
}
\centering
\scalebox{1}{
\begin{tabular}{|c|c|c|c|}
\hline Method & PubMed $\rightarrow$ Reddit1 & PubMed $\rightarrow$ Reddit2 & PubMed $\rightarrow$ Reddit3 \\
\hline
\hline  GCN     & 0.8736 $\pm$ 0.0151 & 0.8996 $\pm$ 0.0158	& 0.8816 $\pm$ 0.0050 \\
\hline  GAT     & 0.9420 $\pm$ 0.0154 & 0.9241 $\pm$ 0.0094	& 0.8985 $\pm$ 0.0088 \\
\hline  DGI     & 0.7845 $\pm$ 0.0208 & 0.9062 $\pm$ 0.0071	& 0.8388 $\pm$ 0.0098 \\
\hline  GPA     & 0.6907 $\pm$ 0.0114 & 0.7146 $\pm$ 0.0050 & 0.7210 $\pm$ 0.0103 \\
\hline UDA-GCN	& 0.9381 $\pm$ 0.0064 & 0.9460 $\pm$ 0.0335	& 0.9506 $\pm$ 0.0099 \\
\hline GPT-GNN	& 0.9582 $\pm$ 0.0087 & 0.9546 $\pm$ 0.0060	& 0.9553 $\pm$ 0.0063 \\
\hline Pretrain-GNN & 0.9405 $\pm$ 0.0059 & 0.9625 $\pm$ 0.0131	& 0.9566 $\pm$ 0.0088 \\
\hline  \name-V & 0.9574 $\pm$ 0.0129 & 0.9761 $\pm$ 0.0059	& 0.9454 $\pm$ 0.0071 \\
\hline  \name-C & \textbf{0.9602 $\pm$ 0.0028} & 0.9749 $\pm$ 0.0043	& 0.9734 $\pm$ 0.0104 \\
\hline  \name   & 0.9575 $\pm$ 0.0072 & \textbf{0.9839 $\pm$ 0.0025}	& \textbf{0.9785 $\pm$ 0.0050} \\
\hline
\end{tabular}}
\label{TB:Single_graph_pubmed}
\end{table*}

\begin{table*} [htp]
\small
\caption{Accuracy of node classification in the setting of multi-source graph transfer, from multiple source graphs (Cora, CiteSeer, and PubMed) to a single target graph (Reddit1 or  Reddit2 or  Reddit3).}
\begin{center}
\scalebox{1}{
\begin{tabular}{|c|c|c|c|}
\hline Method & Reddit1 &  Reddit2 &  Reddit3 \\
\hline
\hline  GCN     & 0.8736 $\pm$ 0.0151 & 0.8996 $\pm$ 0.0158 & 0.8816 $\pm$ 0.0050 \\
\hline  GAT     & 0.9420 $\pm$ 0.0154 & 0.9241 $\pm$ 0.0094 & 0.8985 $\pm$ 0.0088 \\
\hline  DGI     & 0.7845 $\pm$ 0.0208 & 0.9062 $\pm$ 0.0071 & 0.8388 $\pm$ 0.0098 \\
\hline  GPA     & 0.7053 $\pm$ 0.0091 & 0.7193 $\pm$ 0.0027 & 0.7308 $\pm$ 0.0033 \\
\hline  \name-V & 0.9586 $\pm$ 0.0054 & 0.9848 $\pm$ 0.0033 & 0.9801 $\pm$ 0.0024 \\
\hline  \name-C & \textbf{0.9634 $\pm$ 0.0055} & 0.9844 $\pm$ 0.0045 & 0.9795 $\pm$ 0.0051 \\
\hline  \name   & 0.9621 $\pm$ 0.0015 & \textbf{0.9857 $\pm$ 0.0020} & \textbf{0.9811 $\pm$ 0.0025} \\
\hline
\end{tabular}}
\end{center}
\label{TB:Multiple_graph}
\end{table*}

\subsection{Single-Source Graph Transfer}
In this subsection, we first consider the single-source graph transfer setting (following~\cite{DBLP:conf/nips/HuXQYC0T20}), where we are given a single source graph (\eg, Cora~\cite{DBLP:conf/icml/LuG03}, CiteSeer~\cite{DBLP:conf/icml/LuG03}, and PubMed~\cite{namata2012query}) and a single target graph (\eg, Reddit1, Reddit 2, Reddit 3~\cite{hamilton2017loyalty}). Our goal is to pre-train GNNs across two graphs and then fine-tune the model to perform node classification on the target graph. We split the data set into training, validation, and test sets with the fixed ratio of 4\%:16\%:80\%, respectively. Each experiment is repeated five times, and we report the average accuracy and the standard deviation of all methods across different data sets in Tables~\ref{TB:Single_graph_cora} -~\ref{TB:Single_graph_pubmed}.
Results reveal that our proposed method and its variants outperform all baselines. Specifically, we have the following observations: 
(1) compared with the traditional GNNs, \eg, GCN, GAT and DGI, our method and its variants can further boost the performance by utilizing the knowledge learned from both the source graph and the target graph; 
(2) the performance of GPA
is worse than GCN, GAT and DGI. 
One conjecture is that the performance of GPA largely suffers from the label scarcity issue in our setting. In particular, the results of GPA in~\cite{DBLP:conf/nips/HuXQYC0T20} require 67\% labeled samples for training on Reddit1, while we are only given 4\% labeled samples of Reddit1 for training. 
(3) UDA-GCN, GPT-GNN and Pretrain-GNN outperform the traditional GNNs in most cases, which implies that these pre-training algorithms indeed transfer the useful knowledge from the source graph and thus boost the performance of node classification in the target graph;
(4) compared with the pre-training algorithms (UDA-GCN, GPT-GNN and Pretrain-GNN), \name~ further boosts the performance by more than 3\% in the setting of CiteSeer $\rightarrow$ Reddit2.
(5) compared with \name-V which discards the link signal, \name~could further improve the performance by up to 3.31\% by utilizing the structural information of the graph; 
and (6) compared with \name-C which does not utilize the curriculum learning, \name~ boosts the performance by up to 1.88\% in the setting of CiteSeer $\rightarrow$ Reddit2. Overall, the comparison experiments verify the fact that \name\  largely alleviates the impact of negative transfer and improves the generalization performance across all data sets.

\subsection{Multi-Source Graph Transfer}
In this subsection, we evaluate our proposed model \name\ in the multi-Source graph transfer setting, where multiple source graphs are given for pre-training GNNs. In our experiments, we use three source citation networks (\ie, Cora, CiteSeer, and PubMed) as the source graphs, and our goal is to transfer the knowledge extracted from multiple source graphs to improve the performance of the node classification task on a single target graph (\eg, Reddit1, Reddit 2, Reddit 3).
Similar to the single-source graph transfer setting, we use the same data split scheme to generate the training set, the validation set, and the test set. The full results in terms of prediction accuracy and standard deviation over five runs are reported in Table~\ref{TB:Multiple_graph}. 
Note that some of our baseline graph pre-training methods (\ie, UDA-GCN, GPT-GNN, Pretrain-GNN) are not designed for the multi-source graph transfer setting. Thus, we exclude them from Table~\ref{TB:Multiple_graph}. 
In general, our proposed method and its variants outperform all the baselines. 
Moreover, by combining the results (Table~\ref{TB:Single_graph_cora} -~\ref{TB:Single_graph_pubmed}) in the single source-graph setting, it is interesting to see that the performances of GPA and \name\ are both slightly improved when we leverage multiple source graphs simultaneously. For instance, when we consider the Reddit1 as the target graph, the accuracy of GPA and our method is improved by 0.42\% and 0.72\% respectively compared with the single-source graph transfer setting (\ie, Cora $\rightarrow$ Reddit1).

\begin{figure}[t]
    \begin{center}
    \begin{tabular}{cc}
    \includegraphics[width=0.43\linewidth]{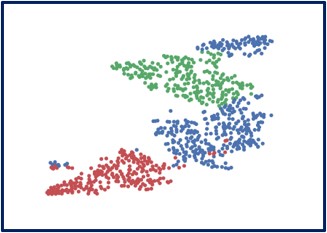} 
    &\includegraphics[width=0.43\linewidth]{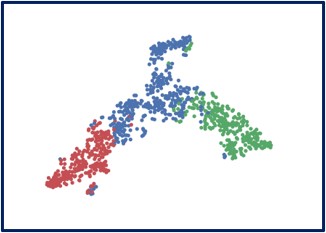}\\
    (a) One-training pair.  & (b)  Two-training pairs. \\
    \includegraphics[width=0.43\linewidth]{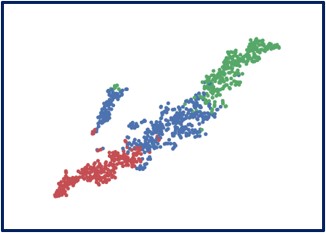}
    &\includegraphics[width=0.43\linewidth]{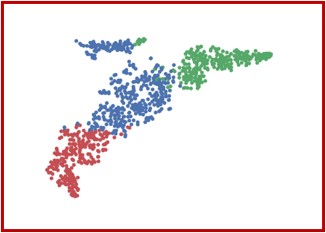}\\
    (c) Three-training pairs. &(d) Ground truth. \\
    \end{tabular}
    \end{center}
    \caption{Metabolic pattern modeling on DPPIN-Breitkreutz graph. (a)-(c) show the network layouts of the generated $\mathcal{G}^5$ by learning from $\{\left \langle\mathcal{G}^1, \mathcal{G}^2\right \rangle\}$, $\{\left \langle\mathcal{G}^1, \mathcal{G}^2\right \rangle, \left \langle\mathcal{G}^2, \mathcal{G}^3\right \rangle\}$, and $\{\left \langle\mathcal{G}^1, \mathcal{G}^2\right \rangle, \left \langle\mathcal{G}^2, \mathcal{G}^3\right \rangle, \left \langle\mathcal{G}^3, \mathcal{G}^4\right \rangle\}$, respectively; and (d) shows the network layout of the ground-truth $\mathcal{G}^5$.}
    \label{Fig: bio_case_study_1}
\end{figure}

\subsection{Case Study: Metabolic Pattern Modeling on Protein-Protein Interaction Graph}
Protein analysis is of great significance in many biological applications~\cite{wang2014dynamic}.
In this case study, we aim to apply \name\ to study and capture the metabolic patterns of yeast cells in the dynamic setting.
To be specific, given a dynamic protein-protein interaction (PPI) graph, e.g., DPPIN-Breitkreutz~\cite{DBLP:journals/corr/abs-2107-02168}, it consists of five snapshots $\mathcal{G} = \{\mathcal{G}^1,\ldots, \mathcal{G}^5\}$. Each node represents a protein, and each timestamped edge stands for the interaction between two proteins. 
In order to train \name, we use the past snapshot as the source graph and the future snapshot as the target graph. 
The five snapshots naturally form four pairs of $\left \langle \text{source graph}, \text{target graph} \right \rangle$ in the chronological order, \ie, $\left \langle\mathcal{G}^1, \mathcal{G}^2\right \rangle$, $\left \langle\mathcal{G}^2, \mathcal{G}^3\right \rangle$, $\left \langle\mathcal{G}^3, \mathcal{G}^4\right \rangle$, $\left \langle\mathcal{G}^4, \mathcal{G}^5\right \rangle$. 
In our implementation, we treat the first three pairs as the training set and use the remaining one for testing. After obtaining the knowledge transfer function $g(\cdot)$ via \name, we aim to reconstruct the last snapshot $\mathcal{G}^5$ by adapting from $\mathcal{G}^4$.  
Figure~\ref{Fig: bio_case_study_1} shows the synthetic $\mathcal{G}^5$ generated by \name\ using different portions of the training set as well as the ground-truth $\mathcal{G}^5$. In detail, Figure~\ref{Fig: bio_case_study_1}(a)-(c) show the generated $\mathcal{G}^5$ by learning from $\{\left \langle\mathcal{G}^1, \mathcal{G}^2\right \rangle\}$, $\{\left \langle\mathcal{G}^1, \mathcal{G}^2\right \rangle, \left \langle\mathcal{G}^2, \mathcal{G}^3\right \rangle\}$, and $\{\left \langle\mathcal{G}^1, \mathcal{G}^2\right \rangle, \left \langle\mathcal{G}^2, \mathcal{G}^3\right \rangle, \left \langle\mathcal{G}^3, \mathcal{G}^4\right \rangle\}$, respectively; and Figure~\ref{Fig: bio_case_study_1}(d) shows the ground-truth $\mathcal{G}^5$ by using t-SNE~\cite{JMLR:v9:vandermaaten08a}. Each dot is the projected node in the embedding space, and the different colors correspond to three substructures of the PPI network. 
In general, we observe that Figure~\ref{Fig: bio_case_study_1}(c) is most similar to the ground-truth in Figure~\ref{Fig: bio_case_study_1}(d), Figure~\ref{Fig: bio_case_study_1}(b) is the next, and Figure~\ref{Fig: bio_case_study_1}(a) is the least similar one. Through this comparison, we know that our \name\ captures the temporal evolution pattern, and the generated molecule graphs are more trustworthy given more temporal information.

\subsection{Parameter Sensitivity Analysis}\label{Parameter Sensitivity}
\begin{figure}[htp]
\begin{tabular}{cc}
\includegraphics[width=0.23\textwidth]{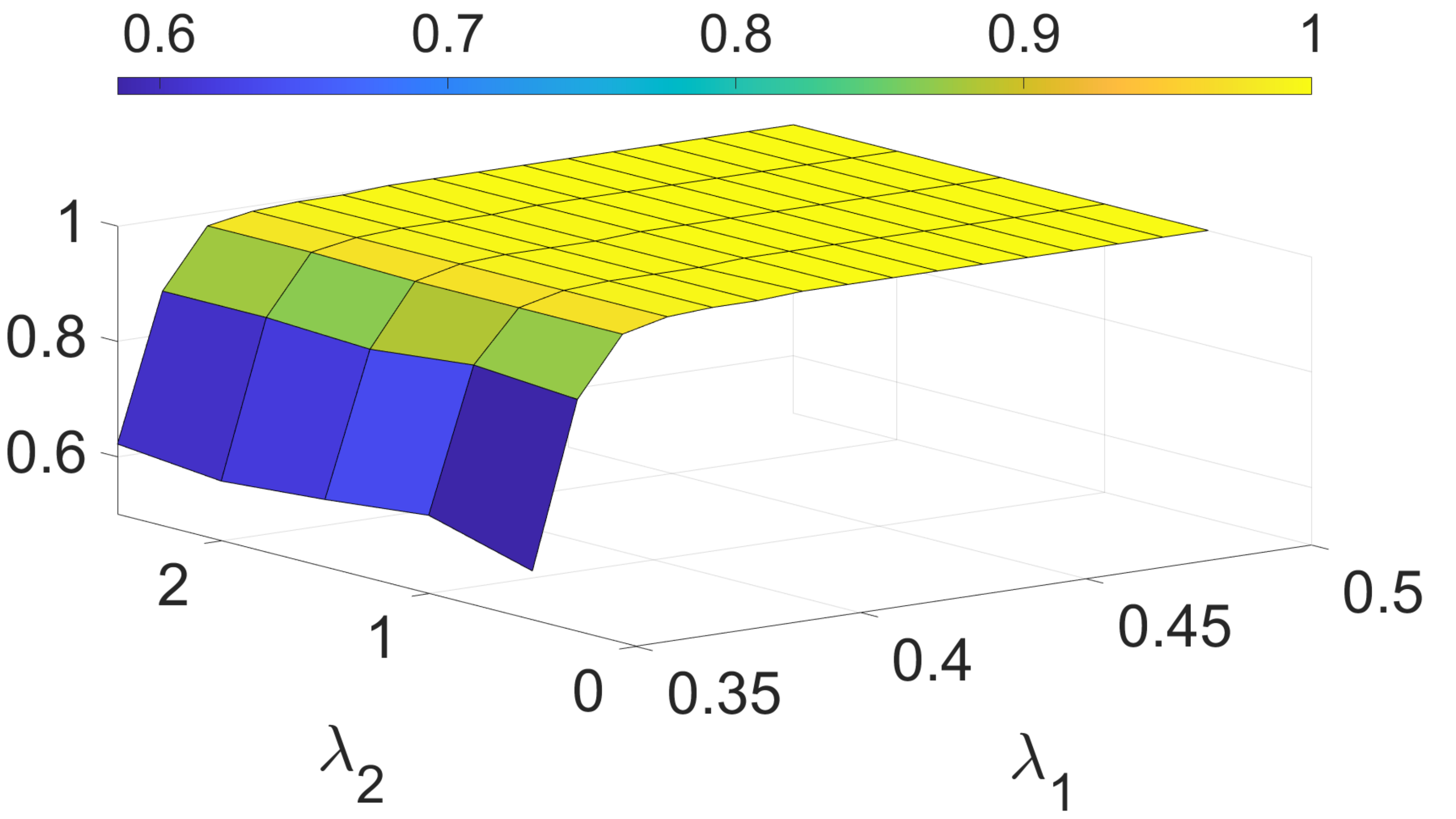} &\includegraphics[width=0.23\textwidth]{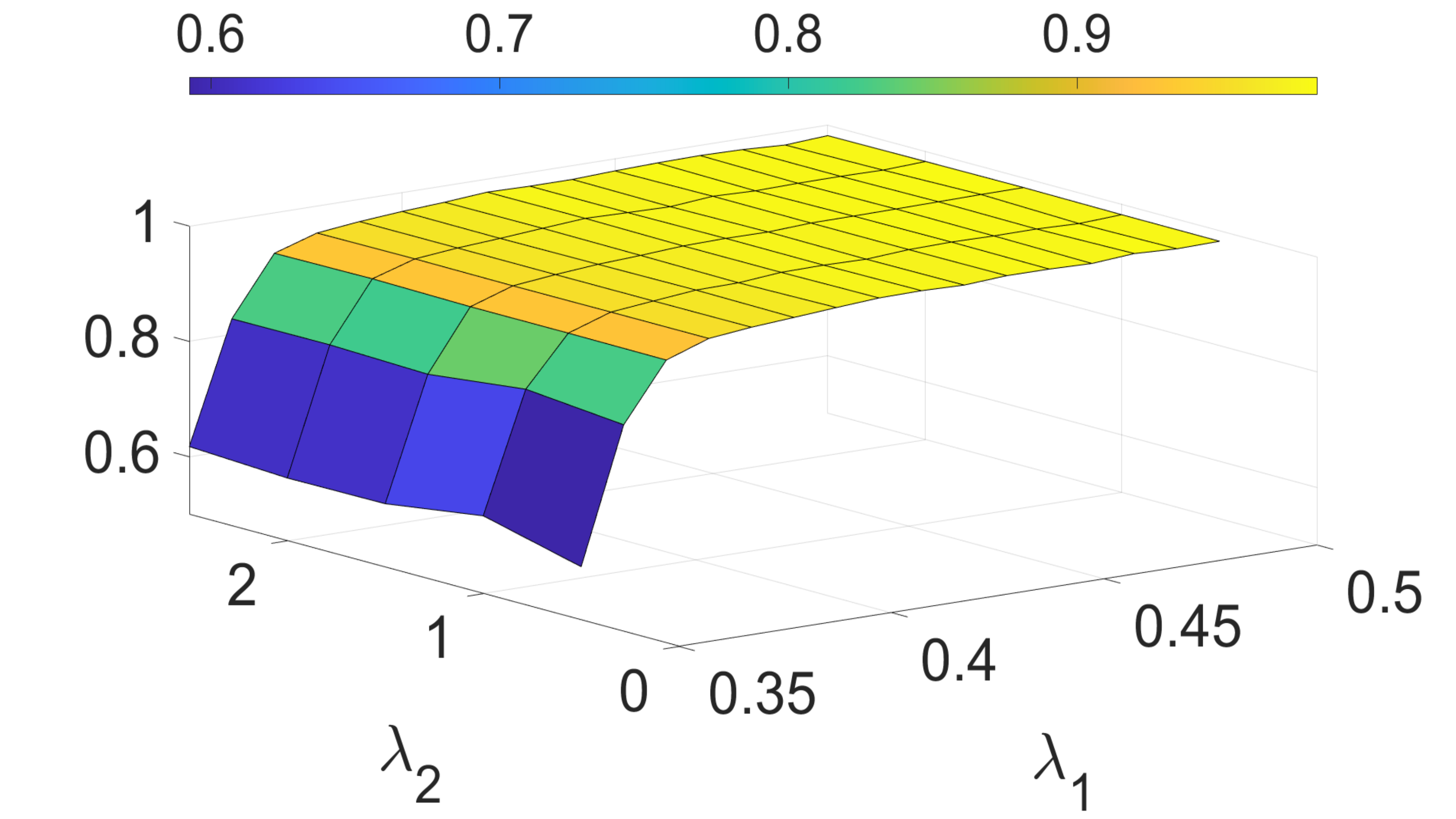} \\
(a) Training accuracy &(b) Testing accuracy \\
\centering
\end{tabular}
\caption{Parameter analysis w.r.t. the learning thresholds $\lambda_1$ and $\lambda_2$ on Cora $\rightarrow$ Reddit2.}
\label{Fig: para}
\end{figure}

In this subsection, we study the impact of the learning thresholds $\lambda_1$ and $\lambda_2$ on \name. 
In Figure~\ref{Fig: para}, we conduct the case study in the single-source graph transfer setting of Cora $\rightarrow$ Reddit2. Particularly, we report the training accuracy and the test accuracy of our model with a diverse range of $\lambda_1$ and $\lambda_2$. In general, we observe that: (1) the model often achieves the best training accuracy and test accuracy when $\lambda_2 =1$; and (2) given a fixed $\lambda_2$, both the training accuracy and test accuracy are improved with increasing $\lambda_1$. It is because the learned teacher model enforces the pre-trained GNNs to learn from the ``easy concepts'' (\ie, small $\lambda_1$) to the ``hard concepts'' (\ie, large $\lambda_1$). In this way, the pre-trained GNNs encode more and more informative graph signals in the learned graph representation and thus achieve better performance.

\subsection{Scalability Analysis}\label{Time Complexity of Pre-Training}
\begin{figure}[t]
\includegraphics[width=0.39\textwidth]{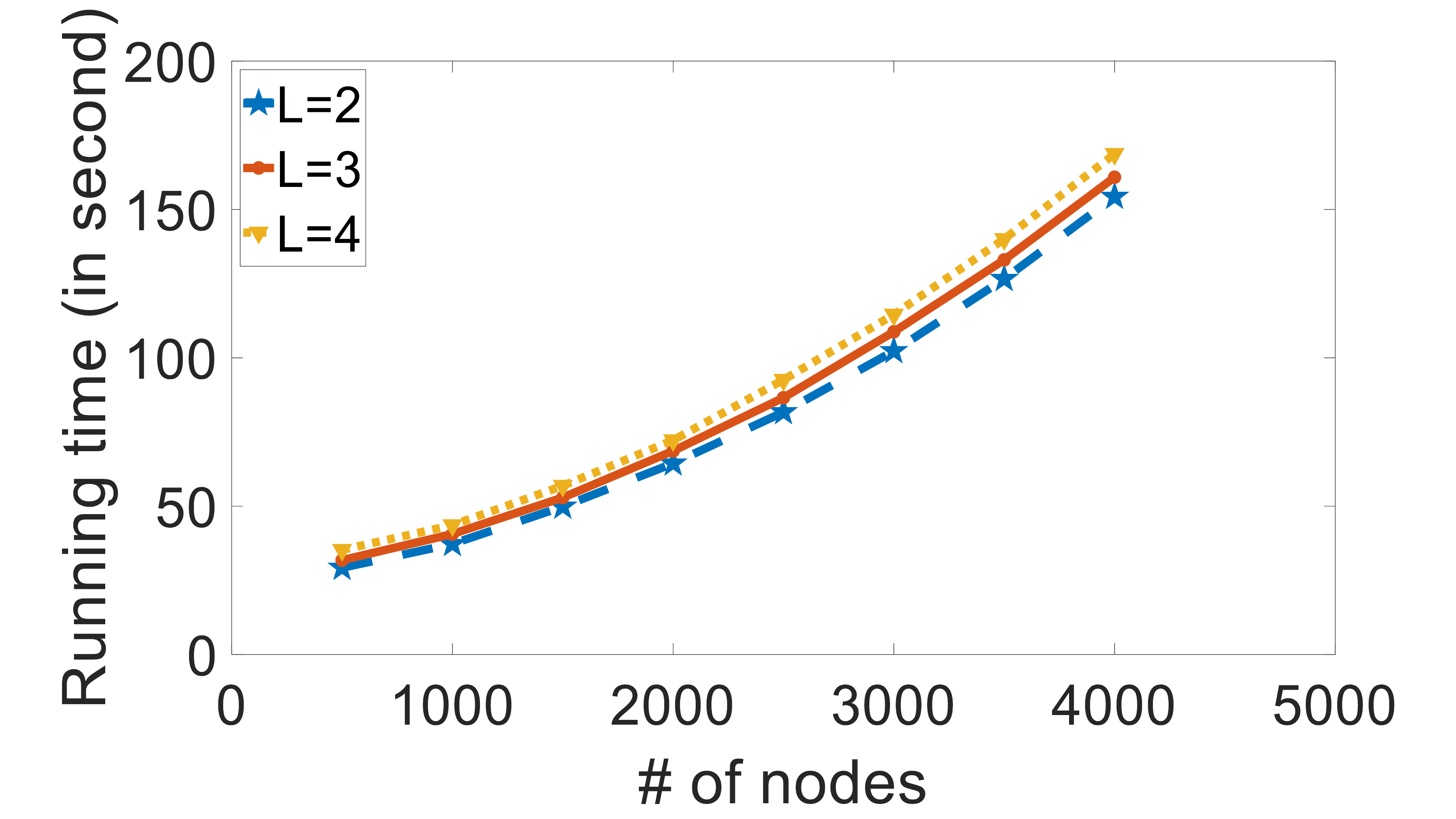}
\centering
\caption{Scalability analysis.}
\label{Fig: Time_complexity}
\end{figure}

In this subsection, we study the scalability of \name\ in the setting of single-source graph transfer, where the source graph is Cora while the target graphs are a collection of synthetic graphs with an increasing number of nodes. 
To control the size of the target graphs, we generate the synthetic target graphs via ER algorithm~\cite{erdos1959random}. 
In Figure~\ref{Fig: Time_complexity}, we present the running time of \name\ with the different number of layers ($L = 2, 3, 4$) over the increasing number of nodes in the target graphs. All the results reported in Figure~\ref{Fig: Time_complexity} are based on 1000 training epochs. In general, we observe that 1) the complexity of the proposed method is roughly quadratic w.r.t. the number of nodes, and 2) the running time only slightly increases when we increase the number of layers in \name.

\section{Related Work}
In this section, we briefly review the related works in the context of pre-training strategies for graphs and domain adaptation.

\textbf{Pre-Training Strategies for Graphs. }
To tackle the label scarcity in graph representation learning, pre-training strategies for GNNs have been proposed~\cite{navarin2018pre, DBLP:conf/iclr/HuLGZLPL20, DBLP:conf/kdd/HuZiniu20,DBLP:conf/kdd/QiuCDZYDWT20, DBLP:conf/aaai/LuJ0S21, DBLP:conf/kdd/JiangJFSLW21,tang2021self, liu2021pre, gritsenko2021graph, jing2021graph}. 
The core idea is to capture the generic graph information across different tasks, and transfer it to the target task to reduce the required amount of labeled data and domain-specific features.
Some recent methods effectively extract knowledge at the level of both individual nodes and the entire graphs via various techniques, including contrastive predictive coding~\cite{DBLP:conf/nips/KhoslaTWSTIMLK20, zheng2021tackling, feng2022adversarial, li2022graph, jing2021hdmi, zheng2021heterogeneous, zheng2022contrastive, jing2022coin}, context prediction~\cite{navarin2018pre,DBLP:conf/kdd/HuZiniu20}, and mutual information maximization~\cite{navarin2018pre,DBLP:conf/iclr/Infograph, DBLP:conf/cikm/JiangL0S21}.
To be specific, 
in~\cite{DBLP:conf/iclr/HuLGZLPL20}, the authors design three topology-based tasks for pre-training GNNs to learn the transferable structural information, (i.e., node-level pre-training, subgraph-level pre-training, graph-level pre-training); 
\cite{DBLP:conf/iclr/HuLGZLPL20} propose a pre-training framework for GNNs by combining the self-supervised pre-training strategy on the node level (i.e., neighborhood prediction and attributes masking) and the supervised pre-training strategy on the graph level (i.e., graph property prediction). 
However, the current pre-training strategies often lack the capability of domain adaptation, thus limiting their ability to leverage valuable information from other data sources. In this paper, we propose a novel method named \name\ that enables pre-training GNNs across graphs and domains. 

\textbf{Domain Adaptation. }
To ensure that the learned knowledge is transferable between the source domains (or tasks) and the target domains (or tasks), many efforts tend to learn the domain-invariant or task-invariant representations, such as~\cite{DBLP:conf/icml/hanzhao2019, DBLP:journals/corr/abs-2010-04647, zhou2020data, zhou2019misc}.
After the domain adaptation demand meets the graph-structured data, the transferability of GNNs has been theoretically analyzed in terms of convolution operations~\cite{DBLP:journals/corr/abs-1907-12972}.
Then to learn the domain-invariant representations with GNNs, several domain-adaptive GNN models are proposed~\cite{DBLP:conf/www/WuP0CZ20, DBLP:conf/icml/LuoWHB20, han2021adaptive, zhu2020transfer, morsing2021supervised}. For example, UDA-GCN is proposed~\cite{DBLP:conf/www/WuP0CZ20} in the unsupervised learning setting to learn invariant representations via a dual graph convolutional network component.
Pragmatically, in order to efficiently label the nodes in a target graph to reduce the annotation cost of training, GPA~\cite{DBLP:conf/nips/HuXQYC0T20} is proposed to learn a transferable policy with reinforcement learning techniques among fully labeled source domain graphs, which can be directly generalized to unlabeled target domain graphs. 
Despite the success of domain adaptation on graphs, little effort has been made to analyze the generalization performance of the deep learning models on relational data (\eg, graphs). Here we propose a new generalization bound for domain-adaptive pre-training on graph-structured data. 

\section{Conclusion}
Pre-training deep learning models is of key importance in many graph mining tasks. 
In this paper, we study a novel problem named domain-adaptive graph pre-training, which aims to pre-train GNNs from diverse graph signals across disparate graphs. 
To address this problem, we present an end-to-end framework named \name, which 1) 
automatically learns a knowledge transfer function and 2) generates a domain-adaptive curriculum for pre-training GNNs from the source graph to the target graph. We also propose a new generalization bound for \name\ in the domain-adaptive pre-training. Extensive empirical results show that the pre-trained GNNs under \name\ with fine-tuning using a few labels achieve significant improvements in the settings of single-source graph transfer and multi-source graph transfer.

\begin{acks}
This work is supported by National Science Foundation under Award No. IIS-1947203, IIS-2117902, IIS-2137468, IIS-19-56151, IIS-17-41317, and IIS 17-04532 and the C3.ai Digital Transformation Institute. The views and conclusions are those of the authors and should not be interpreted as representing the official policies of the funding agencies or the government.
\end{acks}


\balance
\bibliographystyle{ACM-Reference-Format}
\bibliography{sample-base-short}
\end{document}

%% file: math_commands.tex
\usepackage{amsmath,amsfonts,bm}

\def\eqref#1{equation~\ref{#1}}

\def\1{\bm{1}}

\DeclareMathAlphabet{\mathsfit}{\encodingdefault}{\sfdefault}{m}{sl}
\SetMathAlphabet{\mathsfit}{bold}{\encodingdefault}{\sfdefault}{bx}{n}

\DeclareMathOperator*{\argmax}{arg\,max}
\DeclareMathOperator*{\argmin}{arg\,min}